\documentclass[10pt,twocolumn,letterpaper]{article}

\usepackage{wacv}
\usepackage{times}
\usepackage{epsfig}
\usepackage{graphicx}
\usepackage{amsmath}
\usepackage{amssymb}
\usepackage{booktabs}

\usepackage{multirow}
\newcommand{\xdownarrow}[1]{%
  {\left\downarrow\vbox to #1{}\right.\kern-\nulldelimiterspace}
}

\usepackage[table, dvipsnames]{xcolor}

\def\wacvPaperID{1656} 

\wacvalgorithmstrack   

\wacvfinalcopy
\usepackage[pagebackref=true,breaklinks=true,colorlinks,bookmarks=false]{hyperref}

\begin{document}

\title{Local Learning on Transformers via Feature Reconstruction}

\author{Priyank Pathak \quad Jingwei Zhang \quad Dimitris Samaras \\
Stony Brook University, Stony Brook, NY 11794, USA \\
{\tt\small \{prpathak, jingwezhang, samaras \}@cs.stonybrook.edu}
}



\maketitle
\thispagestyle{empty}

\begin{abstract}
Transformers are becoming increasingly popular due to their superior performance over conventional convolutional neural networks(CNNs). 
However, transformers usually require a much larger amount of memory to train than CNNs, which prevents their application in many low resource settings. 
Local learning, which divides the network into several distinct modules and trains them individually, is a promising  alternative to the end-to-end (E2E) training approach to reduce the amount of memory for training and to increase parallelism. This paper is the first to apply Local Learning on transformers for this purpose.
The standard CNN-based local learning method, InfoPro \cite{original}, reconstructs the input images for each module in a CNN. However, reconstructing the entire image does not generalize well.
In this paper, we propose a new mechanism for each local module, where instead of reconstructing the entire image, we reconstruct its input features, generated from previous modules.
We evaluate our approach on 4 commonly used datasets and 3 commonly used decoder structures on Swin-Tiny. 
The experiments show that our approach outperforms InfoPro-Transformer, the InfoPro with Transfomer backbone we introduced,  by at up to $0.58\%$ on CIFAR-10, CIFAR-100, STL-10 and SVHN datasets, while using up to $12\%$ less memory. 
Compared to the E2E approach, we require $36\%$ less GPU memory when the network is divided into 2 modules and  $45\%$ less GPU memory when the network is divided into 4 modules. 

\end{abstract}

\section{Introduction}




Transformers ~\cite{bert, Vaswani} have achieved superior performance in many computer vision tasks such as image classification, segmentation, and detection, and are their usage is rapidly increasing in the computer vision community.
The major advantage of Transformers is their superior performance over conventional convolutional neural networks (CNNs) ~\cite{deit, swinv2, vit}. 
Compared with CNNs, Transformers use local or global self-attention, relaxing the locality constraint introduced by convolutional layers, to improve performance.
However, due to the quadratic complexity of the self-attention mechanism, Transformers require  much larger memory for training compared with convolutional models (CNNs). Under similar conditions and with similar accuracy, DeiT-B Transformer ~\cite{deit} has 2.2 times more parameters than RegNetY-8G ~\cite{RegNetY} (convolutional model). 
This large memory requirement prevents their application in many low resource devices.
Thus there is a strong need for techniques that reduce the memory requirements of the models without sacrificing performance.

The most commonly used paradigm for training deep learning frameworks is end-to-end (E2E) training via backpropagation, where the error gradients computed at the last layer are propagated back to the first layer.
Such a mechanism is memory inefficient as the entire computational graph needs to be stored in the memory with its intermediate variables.
Local learning ~\cite{original, neurongroup, Belilovsky, node_sharing} has been proposed as an alternative to E2E training by dividing a convolutional model into gradient-isolated consecutive modules. 
The primary advantage of local learning is its lower memory cost as it eliminates the need of storing the entire computation graph of the model.
Also, with each module being trained independently, local learning leads to a higher degree of parallelization that further facilitates distributed computing ~\cite{fedrated, fedrated2}. 
This mechanism has also been applied to some real-world applications, such as giga-pixel image classification ~\cite{jingwei} and privacy protection ~\cite{privacy}. 


The major drawback of local learning methods is that they generally achieve inferior performance compared to E2E approaches. 
Recently Wang \etal ~\cite{original} proposed the InfoPro method of local learning on CNNs which encourages the network to preserve as much information as possible with respect to the input image especially in early layers, improving the performance at deeper layers. They achieve this via the reconstruction of input images locally, which brings a significant memory overhead ~\cite{neurongroup}. 



In this paper, we propose a novel local learning framework on Transformers. 
In our framework, for each local module, instead of reconstructing the entire input image like InfoPro ~\cite{original}, we reconstruct the module's input features, generated from previous modules. 
To our best knowledge, this is the first local learning paper on Transformers. 
We compare our method to a baseline that simply integrates Transfomers into the InfoPro framework, which we call InfoPro-Transformer.
Using Swin-Tiny as the backbone, our method outperforms InfoPro-Transformer by up to $0.58\%$ on CIFAR-10, CIFAR-100, STL-10 and SVHN datasets, while using up to $12\%$ less memory. 
Compared to the E2E training strategy, our method requires $36\%$ less GPU memory when the network is divided into 2 modules and requires $45\%$ less GPU memory when the network is divided into 4 modules, at the cost of little or no accuracy drop. 
We further show that our approach works for other variations of the Swin Transformer, namely the Swin-Large and Swin-Base Transformers \cite{swin}.


\section{Related Work}
\begin{figure*}[t]
\begin{center}
\includegraphics[width=1\linewidth]{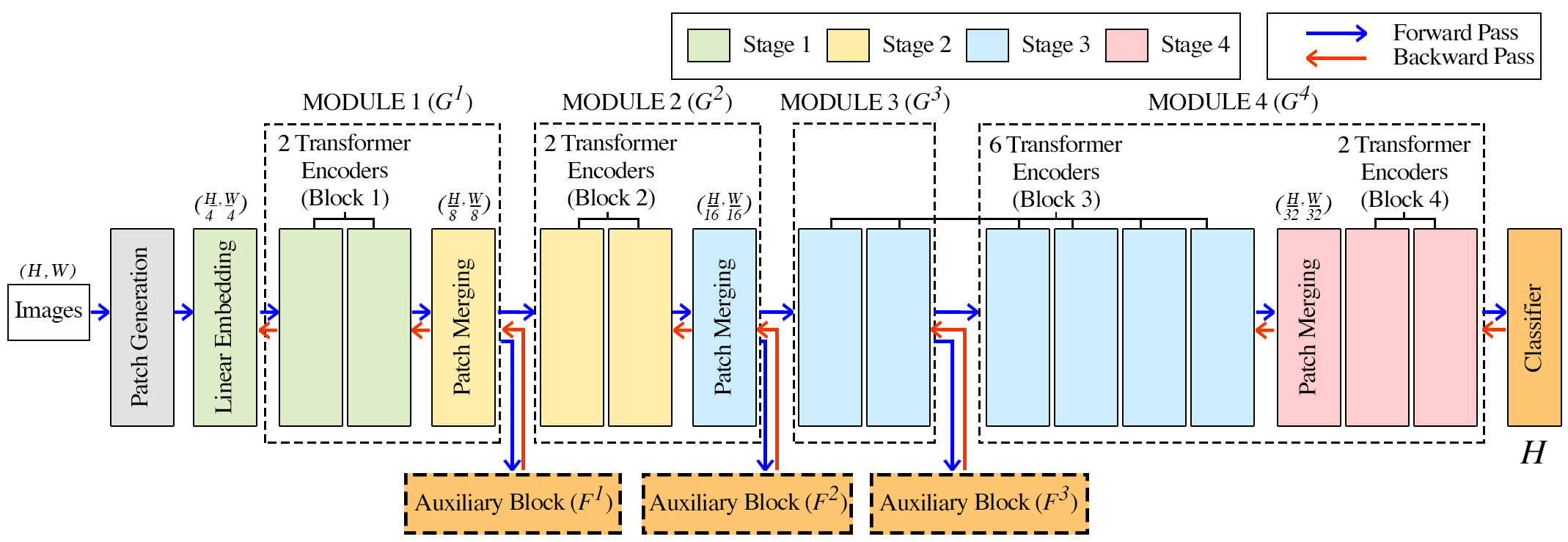}
\end{center}
   \caption{Overview of the proposed local learning framework. Solid lines represent the Swin-Tiny Transformer (Section \ref{sec:swin_arch}). Dotted lines indicate the new Transformer modules (blocks), each trained sequentially and independently from each other (Section \ref{sec:division}). The architecture of the Auxiliary Block is shown in Figure \ref{sec:Auxiliary}. In local learning, gradients (red arrows) are constrained within modules.}
\label{fig:swin}
\end{figure*}

\paragraph{Vision Transformers} Transformers~\cite{Vaswani} were originally proposed to solve natural language problems, and achieved great success in this field with models like BERT ~\cite{bert} and GPT ~\cite{gpt}. 
Vision Transformer (ViT)~\cite{vit} was the first BERT-like Transformer model for computer vision (CV) tasks. 
It tiled the images into patches, embedded them into tokens, and applied multi-head self-attention~\cite{Vaswani} which is a computationally intensive method of introducing the context and long-range dependency on tokens. 
It also employed a trainable positional embedding, which was different from the vanilla Transformer.
Later, Touvron \etal ~\cite{deit} proposed a data-efficient vision Transformer called Deit that has higher throughput and better performance than ViT.
Wang  \etal ~\cite{PVT} proposed Pyramid Vision Transformer (PVT), which downsampled the tokens as the network went deeper and generated features in a pyramid fashion like CNNs, to overcome the difficulties of porting ViT to dense prediction tasks like segmentation. 
Swin Transformer ~\cite{swin}, is the first Transformer model that outperformed the state-of-the-art CNNs at that time on segmentation and detection tasks. They constrained the self-attention within local windows, which allowed shift operation for exchanging context across the windows. 
They used relative positional embedding ~\cite{bao2020unilmv2} and like PVT, also adopt a pyramid structure.
Some recent works ~\cite{CVT, longsformer, xiong2021nystromformer} have successfully brought down the complexity of using various Transformers.
Despite these measures, training Transformers still remains a memory-intensive task. 


\paragraph{Local Learning} 
Local learning ~\cite{Nokland, Belilovsky, greedy1} is an alternative to E2E training to train neural networks. 
It divides  deep learning models into consecutive modules that can be trained  independently. Thus the error gradients are constrained in their specific modules (gradient isolated), via local loss functions.
Some approaches like Cascade Learning ~\cite{cascade1, cascade2, cascade3, Belilovsky2} trained one module fully before moving on to the next module. 
While other approaches ~\cite{Nokland, Belilovsky, greedy1} trained all modules simultaneously. 
In this work, we primarily focus on the second approach as it is more practical, given its ability to be easily parallelized and distributed ~\cite{fedrated, fedrated2}.

%
%
Belilovsky \etal ~\cite{Belilovsky2} attached an additional classifier to each local module to predict the final target and evaluated their method on ImageNet~\cite{imagenet}.
Wang \etal ~\cite{original} showed that employing classification loss (\eg cross entropy) locally discarded task-relevant information (distinguishing features) at early layers/modules and thus hindered the performance of deeper layers. 
They thus proposed InfoPro and achieved a major improvement over previous work. 
To preserve task-relevant information in early modules, InfoPro reconstructed the input images locally at each module and maximized the information gain between the input images and the generated features. 
However, the reconstruction of the entire input images led to high computational cost as suggested by ~\cite{neurongroup}. Du \etal ~\cite{cascade1} argued that maximizing information gain via an auxiliary decoder was biased toward the choice of the auxiliary network. In contrast, to previous approaches, our method relies solely on reconstructing features instead of input images, reducing the complexity significantly. Our approach is similar to \cite{feature1}, which reconstructs features in a residual network for regularization purposes, without any decoder. 

Zhang \etal ~\cite{jingwei} proposed reconstructing patches instead of reconstructing the entire image to reduce the memory consumption for giga-pixel images. 
But they did not provide any analysis on reconstructing the features versus reconstructing the input image.
Guo \etal ~\cite{node_sharing} showed improvements in the performance by sharing convolutional nodes across the modules instead of reconstructing the input, albeit at the cost of a small memory overhead. Sharing a node in Transformer quickly becomes impractical as sharing Transformer blocks (basic units in a Transformer) would have a huge memory overhead. 


\section{Methodology}
\subsection{The structure of Swin Transformer}
\label{sec:swin_arch}
In this paper, we mainly use the  Swin-Tiny architecture as the backbone. As shown in Figure \ref{fig:swin}, colored rectangles with solid lines indicate the layers of a Swin-Tiny Transformer. 
For an input image $x$ with height $H$ and width $W$, the network first converts the input image into patches, grouping $4\times 4$ pixels together (Patch Generation).
Thus the input generates $\frac{H}{4} \times \frac{W}{4}$ number of patches. 
The first layer (Linear Embedding) embeds these patches into tokens (features). 
Then these patches sequentially pass through 4  Transformer stages, each encompassing an even number of Transformer blocks. These blocks occur in a pair, namely W-MSA and SW-MSA, which alternates within a stage. 
Stages are divided according to the number of tokens.
Layers generating the same number of tokens are grouped into the same stages.
As shown in Figure \ref{fig:swin}, green, yellow, light blue and pink rectangles represent the 4 stages respectively.

\subsection{Overview of our framework}
Our local learning framework first divides the entire Transformer network layer by layer into $K$ consecutive modules ($G^1$ , $G^2$, ... $G^K$) and optimizes them separately. Such a network is trained using pairs of $(x, y)$ on which $x$ denotes the input image and $y$ the corresponding label.

A module (split) contains layers of the original network, for example, the first module contains $2$ transformer encoders and a Patch Merging layer in a Swin-T.
We assume the input to a module $G^i, i=1,..,K$ is $x_{i - 1}$ and the output is $x_i = G^i (x_{i-1}) $, and $x_0$ is the input image. 
An overview of our approach is presented in Figure \ref{fig:swin} in which the forward and backward passes are indicated. 

To train each module $G^i$ with its input $x_{i-1}$ locally, we use an Auxiliary Block $F^i$, and compute the loss as $\mathcal{L}_i = F^i(G^i(x_{i - 1}), x_{i-1}, y) = F^i(x_i, x_{i - 1}, y)$. 
Module $G^i$ is trained by minimizing $\mathcal{L}_i$.
Then the next module $G^{i+1}$ has an input $x_i = G^i(x_{i - 1})$ and is trained using the next Auxiliary Block $F^{i+1}$ using the same method as above. 
This method is applied to all local modules except the last one. 
The last local module $G^K$ is trained using only the final classifier $H$. 
The Linear Embedding layer, at the beginning of the network, does not belong to any module and is trained using the gradients passed from the first module.

\subsection{Network division}
\label{sec:division}
We preserve the Swin Transformer structure of alternating blocks within all our local modules, hence we divide the network only after SW-MSA encoder. Patch merging layers are grouped together with their previous Transformer Block for implementation convenience. 
Usually, there are multiple ways to divide a network into $K$ modules.
As shown in Figure \ref{fig:swin}, for a Swin-T network, if we divide it into $4$ modules, the 1st to 4th module (surrounded by dotted lines) has 2,2,2,6  transformer blocks respectively.
The division of other variations of Swin Transformer, Swin-Base and Swin-Large is shown in Table \ref{tab:configuration}. 
We have derived these configurations empirically based on memory footprints of each possible way of splitting the model. We chose the configuration that minimizes the maximum amount of computational memory of the local modules.


\begin{table}
  \begin{center}
    {\small{
\begin{tabular}{ccccc}
\toprule
Swin type & Configuration & K=2 &  K=4 \\
\midrule
Swin-Tiny & 2,2,6,2 & 2,10 &  2,2,2,6\\
Swin-Base  & 2,2,18,2 & 8,16 &  2,6,8,8\\
Swin-Large  & 2,2,18,2 & 8,16 & 2,6,8,8\\
\bottomrule
\end{tabular}
}}
\end{center}
\caption{Configuration of Swin-Tiny and Swin-Large means the number of Transformer blocks in each stage. $K=2$ and $K=4$ indicate the distribution of Transformer blocks in each module (Section \ref{sec:division}) }
\label{tab:configuration}
\end{table}


\subsection{Auxiliary Block}
\label{sec:Auxiliary}

\begin{figure}[t]
\begin{center}
   \includegraphics[width=1\linewidth]{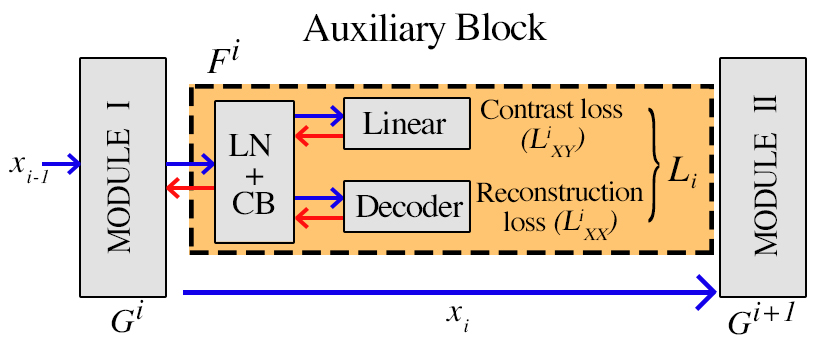}
\end{center}
   \caption{Auxiliary Block architecture (Section \ref{sec:Auxiliary}). $G^i$ is the "$i$-th" module with an input $x_{i-1}$. The output is forwarded to the the Auxiliary Block and next module $G^{i+1}$. The Blue arrow represents forward pass, while the red arrow indicates error gradients. LN refers to layer normalization, CB refers to convolutional block, and MLP is a linear layer.}
\label{fig:auxiliary}
\end{figure}

Our overall design of Auxiliary Block $F^i$ is similar to that of InfoPro. 
As shown in Figure \ref{fig:auxiliary}, input feature $x_{i-1}$ is passed to the $i$-th module $G^i$ which outputs $x_i=G^i(x_{i-1})$.
$x_i$ is passed to the auxiliary block $F^i$ (the orange rectangle). 
The Auxiliary Block generates a local loss $L_i$ to supervise the local module $G^i$.
Inside the block, $x_i$ is first passed to a network consisting of Layer Normalization (LN) followed by a convolutional block for dimensionality reduction (denoted as ``LN + CB"). The output  is then passed to two branches, one for contrastive loss and the other for reconstruction loss.

In the contrastive loss branch, the input is passed to a fully connected layer (linear layer), to generate a feature vector. 
This feature vector and its corresponding label $y$ are used to generate a contrastive loss $L^i_{XY}$. 
Chen \etal \cite{contrast_loss} proposed the NT-Xent contrastive loss, to maximize the similarity between embeddings of samples that have undergone augmentations. InfoPro repurposed their loss for local learning and we do the same here. The idea is to increase the similarity between embeddings of samples taken from the same class.

\begin{figure}
\begin{center}
   \includegraphics[width=1\linewidth]{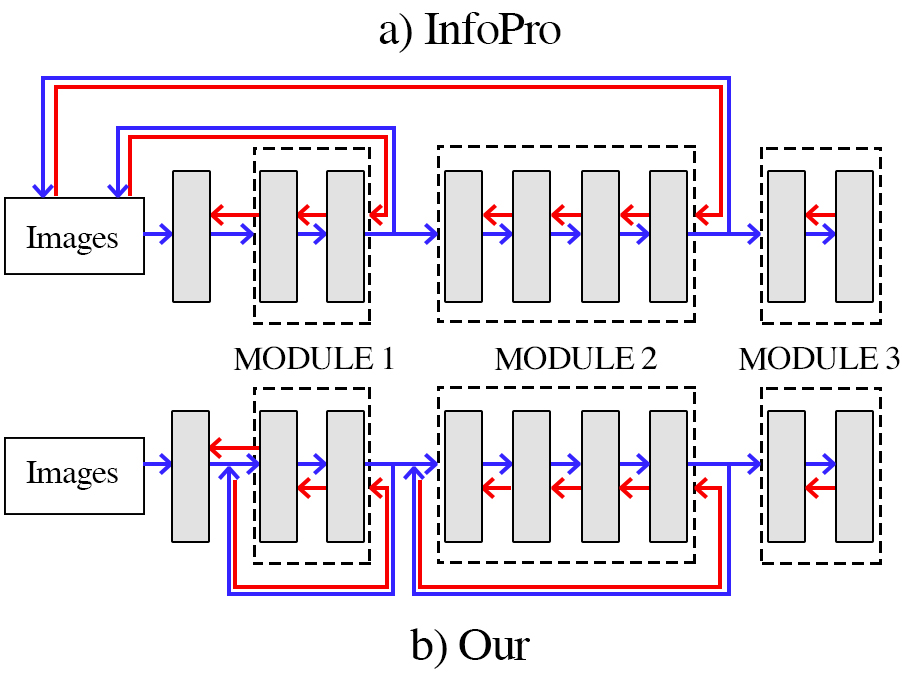}
\end{center}
   \caption{(a) Reconstructing images (InfoPro) vs (b) reconstructing features (Ours). }
\label{fig:Difference}
\end{figure}

In the reconstruction loss branch, the input is passed to a decoder to generate a feature map trying to reconstruct feature $x_{i-1}$. 
The reconstruction loss $L^i_{XX}$ is generated between the feature map and $x_{i-1}$ to measure their similarity.
Figure \ref{fig:Difference} shows the difference between the reconstruction of our method and that of the InfoPro approach. 
InfoPro, shown in a), reconstructs the input image multiple times using the features of each module. Our method, is shown in b) which reconstructs the input features of the modules. 
Note for the first module, we reconstruct the input patch embeddings instead of the original image $x_0$.
Feature maps usually have smaller spatial dimensionality, hence the upscale factor is smaller compared to the input image, which makes our method require less GPU memory than InfoPro.

\subsection{Optimization}


The first $K-1$ modules are optimized locally with the following setting:
\begin{align}
    \mathcal{L}_i &= F^i  \Bigl( G^i(x_{i - 1}), x_{i-1}, y \Bigr) \\
    &= \alpha_i L_{XX}^i + \beta_i L_{XY}^i 
\end{align}
where $i$ represents the $i$-th module and hyperparameters $\alpha_i$,$\beta_i$ are
regularization terms. $\alpha_i$ represents the emphasis on reconstruction, while $\beta_i$ represents the weighting of target labels in shaping the features.  
As we go deeper into the model, $i=1,..,K-1$,  $\alpha_i$ decreases linearly while $\beta_i$ increases linearly.
We use Binary Cross Entropy (BCE) for the reconstruction loss $L_{XX}^i$ and NT-Xent \cite{original, contrast_loss} for the constractive loss $L_{XY}^i$.
For the last local module $G^K$, the training scheme is changed to
\begin{align}
    \mathcal{L}_K &= CE(H(G^K(x_{K-1})), y)
\end{align}
where $H$ represents  the final classifier and $CE$ represents the Cross Entropy loss.

\section{Experiments}
\begin{table*}
\begin{center}
    \begin{tabular}{c  c  c  r  r  r  r c } 
    \toprule
        \multicolumn{1}{c}{Split} & \multicolumn{1}{c}{Decoder} & \multicolumn{1}{c}{Method} & \multicolumn{1}{c}{CIFAR-10} & \multicolumn{1}{c}{CIFAR-100} & \multicolumn{1}{c}{STL-10} & \multicolumn{1}{c}{SVHN} & \multicolumn{1}{c}{Memory (MB) }\\
    \midrule
    - & - & E2E & \underline{98.43} & \underline{88.99} & \underline{98.72} & 97.52 & 14769 \\ 
    \midrule

    & & InfoPro  & 98.12 & \textbf{88.39} & \textbf{98.51} & $\underline{\textbf{98.12}}$ & 9557 \\ 
    \rowcolor{lightgray!60}\cellcolor{white} & \cellcolor{white} \multirow{-2}{*}{Interp.} & Ours & \textbf{98.13} & 88.28 & \textbf{98.51} & 97.30 & 9390 \color{red} $(\downarrow 1.74\%)$\\  
     
    \cline{2-8} 
    
    & & InfoPro  & 98.00 & 87.88 & \textbf{98.46} & 97.29 & 9456 \\ 
    \rowcolor{lightgray!60}\cellcolor{white} & \cellcolor{white} \multirow{-2}{*}{Linear} & Ours & \textbf{98.20} & \textbf{87.90} & 98.35 & \textbf{97.31} & 9388 \color{red} $(\downarrow 0.72\%)$\\  
    
    \cline{2-8} 
    & & InfoPro  & 98.07 & 87.83 & \textbf{98.46} & 96.85 & 9458 \\ 
    \rowcolor{lightgray!60}\cellcolor{white} \multirow{-6}{*}{K=2} & \cellcolor{white} \multirow{-2}{*}{Deconv.} & Ours & \textbf{98.30} & \textbf{87.91} & \textbf{98.46} & \textbf{97.25} & 9390 \color{red} $(\downarrow 0.72\%)$\\ 
  
    \midrule
    
    &  & InfoPro & 95.89 & 75.59 & 97.65 & \textbf{95.39} & 8076  \\ 
    \rowcolor{lightgray!60}\cellcolor{white} & \cellcolor{white} \multirow{-2}{*}{Linear} & Ours & \textbf{96.47} & \textbf{78.56} & \textbf{97.66} & 95.03 & 8037 \color{red} $(\downarrow 0.48\%)$ \\ 
    
    \cline{2-8}  
    & & InfoPro & \textbf{96.55} & 79.93 & 97.68 & \textbf{95.40} & 9204   \\ 
\rowcolor{lightgray!60}\cellcolor{white} \multirow{-5}{*}{K=4} &  \cellcolor{white} \multirow{-2}{*}{Deconv.}  
& Ours & 96.51 & \textbf{80.08} & \textbf{97.93} & 95.21 & 8086 \color{red}$(\downarrow 12.15\%)$\\ 

    \bottomrule
    \end{tabular}
\caption{
Comparison of accuracy and GPU memory consumption for ImageNet pretrained Swin-Tiny, finetuned on 4 datasets. We compare InfoPro and our approach using three decoders, with \textbf{Bold} indicating the superior performance among the two. The \underline{underline} indicates the overall best performance. Interp. is a bilinear interpolation based decoder, Linear is a fully connected layer decoder, and Deconv. is stacked deconvolution layers. The red percentage next to memory indicates the percentage of memory saving our approach offers over InfoPro. MB refers to megabyte of GPU memory occupied by the model. }
\label{tab:pretrained}
\end{center}
\end{table*}

\subsection{Datasets}
We evaluate our model’s performance on 4 commonly used datasets, namely CIFAR-10, CIFAR-100 ~\cite{cifar10}, SVHN ~\cite{svhn}, and STL-10 ~\cite{stl10}. CIFAR-10 and CIFAR-100 both contain $50,000$ training images and $10,000$ test images, with $10$ and $100$ target labels, at $32 \times 32$ resolution respectively. STL-10 has $5000$ training images with $10$ classes and, $8000$ test images at $96 \times 96$ resolution. SVHN also has $10$ classes, with $73,257$ training images with $26,032$ test images at $32 \times 32$ resolution. We scale up the input image sizes to $224 \times 224$. 

\subsection{Implementation Details}
Unless explicitly mentioned, we use Swin-Tiny for all our experiments. Our training pipeline is identical to Swin Transformers ~\cite{swin}, consisting of the AdamW optimizer with a weight decay of $0.05$, base learning rate of $0.0005$, and a cosine decay learning rate scheduler. 
We run our experiments on four NVIDIA TITAN RTX or three Quadro RTX 8000.
The memory consumption is measured via the Pytorch method $max\_memory\_allocated$ as performed by \cite{node_sharing}.
During the evaluation, the network is treated as E2E and the input is passed from the first layer to the last layer with the auxiliary blocks not playing any role. During training, we have two loss hyperparameters, ($\alpha_i$,$\beta_i$), whose values range between $(3.0, 1.0), (2.0, 2.0), (1.0, 3.0), (0.0, 4.0)$ as we go deeper in the model. For $K=2$, ($\alpha_i$,$\beta_i$), range between $(3.0, 1.0), (0.0, 4.0)$.

\subsection{Results}
We choose overall accuracy (Top-1 Accuracy) as the main metric for the evaluation of our method and we perform an extensive study to compare our method with InfoPro-Transformer method. 
InfoPro-Transformer is the method we introduce by replacing the CNN backbone in InfoPro to Swin-Tiny for fair comparison with our method.
We use InfoPro to represent InfoPro-Transformer for simplicity.
As the performance of our method and InfoPro method may depend on the structure of decoder, to prove the superior performance of our method, we further evaluate our method and InfoPro using three commonly used structures of decoders. 
The first one is a combination of bilinear interpolation and CNNs, which is the structure that InfoPro used in their experiments. We use Interp. to represent it. For $224 \times 224$ input (our case), InfoPro provide details for only $K=2$, hence we limit its ussage only for $K=2$. The second one is simple a fully connected linear layer, which is used in \cite{simm}. We use Linear to represent it. The third one is a simple stack of deconvolution layers \cite{stack_deconv}, with a upscaling factor of $2\times2$ and $4 \times 4$. We use Deconv. to represent it.

\subsubsection{Accuracy comparison using pre-trained network}
\label{sec:pretrained_acc}

We first compare our method with InfoPro when the network is pretrained on ImageNet \cite{imagenet}, the most common scenario for image classifications. 
The four datasets are all trained for 50 epochs with batch size 140. 
The results are shown in Table \ref{tab:pretrained}, for $K = 2$, on the CIFAR-10 dataset, our method surpasses InfoPro by $0.01\%$, $0.20\%$, and $0.23\%$ on all three decoders. 
For $K = 4$, our method, on the CIFAR-10 dataset, achieves $0.58\%$ higher accuracy when using Linear decoder.
This number looks small but given all methods statute on this dataset, this improvement is significant. 
On the other $3$ datasets, we have very similar results, for $K = 2$ and $K = 4$ our method is comparable to the InfoPro.

We also compare our method with end-to-end (E2E) training. As shown in Table \ref{tab:pretrained}, our method of $K = 2$, whatever decoder are we using, achieves comparable accuracy with E2E ones.

\begin{table*}[t!]
\begin{center}
    \begin{tabular}{c  c  c  c  c  c  } 
    \toprule
        \multicolumn{1}{c}{Split} & \multicolumn{1}{c}{Decoder} & \multicolumn{1}{c}{Method} & \multicolumn{1}{c}{CIFAR-10} & \multicolumn{1}{c}{CIFAR-100} & \multicolumn{1}{c}{STL-10} \\
    \midrule
    - & - & E2E & \underline{91.37} & 75.03 & 72.19    \\ 
    \midrule
    
    &  & InfoPro  & \textbf{90.74} & 71.72 & 74.61   \\ 
    \rowcolor{lightgray!60} \cellcolor{white} 
    & \cellcolor{white} \multirow{-2}{*}{Interp.} & Ours  & 90.55 & \textbf{72.74} &  \textbf{74.89}   \\ 
    \cline{2-6}

    &  & InfoPro &  90.55  & 72.81 & 74.62  \\ 
    \rowcolor{lightgray!60}\cellcolor{white}
    & \cellcolor{white} \multirow{-2}{*}{Linear} & Ours & \textbf{90.79}  & \textbf{73.64} & \textbf{74.89}   \\ 
    \cline{2-6}
    
    &  & InfoPro  & 90.85 & 72.84  &  74.63   \\ 
    \rowcolor{lightgray!60}\cellcolor{white}
    \multirow{-6}{*}{K=2} & \cellcolor{white} \multirow{-2}{*}{Deconv.} & Ours &  \textbf{91.18} &  \underline{\textbf{75.72}} & \underline{\textbf{75.31}} \\ 
    
    \midrule
    
    
    &  & InfoPro  &  81.22 & 68.36 & 64.61  \\ 
    \rowcolor{lightgray!60}\cellcolor{white}
    & \cellcolor{white} \multirow{-2}{*}{Linear} & Ours & \textbf{84.80}  & \textbf{72.04} & \textbf{64.80}  \\ 
    \cline{2-6}
    
    && InfoPro & 83.38 & 64.09 & 58.69   \\ 
  \cline{3-6}
    \rowcolor{lightgray!60}\cellcolor{white}
    \multirow{-6}{*}{K=4} & \cellcolor{white} \multirow{-2}{*}{Deconv.}  & Ours & \textbf{86.28} & \textbf{73.00}  & \textbf{66.89} \\ 
    \bottomrule
    \end{tabular}
\caption{Comparison of accuracy for Swin-Tiny trained from scratch. For a given decoder, \textbf{Bold} indicates the superior performance between our approach and InfoPro, while \underline{underline} indicates the overall best performance.}
\label{tab:scratch}
\end{center}
\end{table*}

\subsubsection{Evaluation of GPU memory consumption}

A major advantage of local learning is that, compared with E2E training, it saves a significant amount of GPU memory.
As shown in Table \ref{tab:pretrained}, when diving the network into 2 modules ($K=2$), our method, whatever the decoder is, requires only less than 9.4 GB of memory. 
Compared with E2E training, which consumes around 14.8 GB of memory, our method reduces memory consumption by $36\%$.
This memory reduction is further enlarged to around $45\%$ when the network is divided into 4 modules ($K=4$), with only a little sacrifice on the accuracy.

We also compare our method with InfoPro on memory consumption. 
Our method (reconstructs features) is relatively cheaper than the InfoPro (reconstructs the entire images). 
This is because, for our Swin-Tiny configuration, reconstructing the image of resolution $224 \times 224$ is a costly operation with a minimum upscale factor of 64 ($(\frac{H}{8},\frac{W}{8}) \rightarrow (H,W)$ for $F^1$ in Figure \ref{fig:swin}), while reconstructing features has a maximum upscale factor of 4 (\eg $(\frac{H}{8},\frac{W}{8}) \rightarrow (\frac{H}{4},\frac{W}{4})$ for $F^1$). As shown in Table \ref{tab:pretrained}, for $K = 2$ and $K = 4$ our method saves slightly more memory than the InfoPro method in most cases.
For the $K=4$ case using Deconv. decoder, our method largely reduces (by $12.15\%$) the $9204$ MB GPU memory requirement of InfoPro and only requires $8086$ MB of memory.

For $K = 2$, a similar memory footprint of three decoders for feature reconstruction indicates a similar memory overhead of three decoders. For $K = 4$, multiple deconvolutions are stacked to create an upscale factor of $16\times16$ for image reconstruction, hence memory overhead is abnormally for InfoPro deconvolutional (Deconv.) decoder.


\subsubsection{Accuracy comparison using random-initialized network}
\label{sec:scratch}
Despite our comparison using ImageNet pre-trained networks in section \ref{sec:pretrained_acc}, we further evaluate our method using random-initialized networks on three datasets. 
All the models are with a batch size of 140. 
All experiments on STL-10 are trained for 300 epochs and all experiments on CIFAR-10 and CIFAR-100 are trained for 200 epochs. Due to the enormous training cost and following common practice \cite{small2, small1}, we evaluate STL-10 at 300 epochs and evaluate CIFAR-10 and CIFAR-100 at 200 epochs. Note that all other training settings remain the same across all experiments.
As presented in Table \ref{tab:scratch}, our method consistently outperforms InfoPro on three decoders for both $K = 2$ and $K = 4$. 

When the network is divided into $2$ modules $(K = 2)$, our method generally outperforms InfoPro except for a comparable performance for CIFAR-10 using the Interp. decoder. Note that our method $(K = 2)$ uses a Deconv. decoder surpasses end-to-end (E2E) model for CIFAR-100 and STL-10 by $0.69\%$ and $2.78\%$ respectively. When the network is divided into $4$ modules $(K = 4)$, our method outperforms InfoPro by a much larger margin than that of $K = 2$. Our method uses Deconv. decoder on CIFAR-100, achieves $73.03\%$ accuracy, which is $8.98\%$ better than the InfoPro method. It is the largest improvement in our experiments.



\subsubsection{Ablation on losses}

As shown in section \ref{sec:scratch}, in some experiments, our local learning method surpasses the end-to-end trained model, which is counter-intuitive.
We hypothesize that the contrastive loss, which the E2E training pipeline do not have, is the reason of such superior performance.
We then conduct an ablation study on the losses to prove our hypothesis. 
We compare our approach using Deconv. deocer on CIFAR-100 with two greedy approaches that do not have the reconstruction branch in auxiliary blocks.
One greedy approach is our method without the the reconstruction branch.
The other greedy approach replaces contrast loss with cross entropy loss. 
This approach is similar to E2E as it only has several cross entropy losses.

As shown in Table \ref{tab:greedy_approach}, the greedy approach with cross entropy loss achieves $69.11\%$ accuracy, is $5.92\%$ lower than the E2E network.
While the greedy approach with constrictive loss achieves $75.64\%$ accuracy, $0.61\%$ higher than the E2E network.
It shows that the major performance gain is because of the constrictive loss, proving our assumption.
Also, our model combines the advantages of contrastive learning and our , thus amplies the performance even further as indicated 0.08\% gain in the accuracy. 

\begin{table}
\begin{center}
    \begin{tabular}{c  l  c  c} 
    \toprule
    Split & Method & Acc. & Mem. (MB)  \\ 
    \midrule
    - & E2E &  75.03  & 14769  \\ 
    \midrule
    \multirow{3}{*}{K=2} & Greedy (Cross entropy) &  69.11 & 9378  \\
    & Greedy (Contrast) &  75.64  & 9378  \\
    & Ours (Deconv.) &  75.72 &  9390 \\
    \bottomrule
    \end{tabular}
\caption{Ablation on losses. We compare our approach using Deconv. decoder with greedy approach of training without any reconstruction. (Contrast) uses only Contrast loss at local modules, while (Cross entropy) uses cross entropy loss only. }
\label{tab:greedy_approach}
\end{center}
\end{table}

\subsubsection{Variations of Swin}

In order to show our methodology can for work for all types of transformers, we pick two of variations of Swin Transformers, namely Swin-Base and Swin-Large without any pre-training.
Swin-Base is trained for on 3 GPU with a batch size of 40 on an input image resolution of $384 \times 384$. Because of excessive compute and training costs, we restrict ourselves with 100 epochs. 
Table \ref{tab:Base_model} shows a significant saving on GPU memory with $K=2$ and $K=4$ saving around $43.81 \%$ and $66.59\% $ respectively. On STL-10 we see a gain in performance for $K=2$, while $K=4$ is comparable to E2E. For CIFAR-100, we observe an interesting phenomenon, where $K=4$ surpasses E2E performance. 

Swin-Large results are shown in Table \ref{tab:large_model}. Here also we face serious compute limitations, hence limit the training with an input resolution of $224 \times 224$, for a batch size of 80 trained for 200 epochs. Here we observe the memory saving is less compared to Swin-Base, but its still significantly greater than Swin-Tiny, saving around $35.31\%$ and $60.11\%$ GPU memory for $K=2$ and $K=4$ respectively. On CIFAR-100, $K=4$ results surpasses E2E by a margin of $19.47\%$ suggesting a faster rate of convergence.

In summary, we save significant amount of memory with our technique surpassing E2E for split $K=4$.  
\begin{table}
\begin{center}
    \begin{tabular}{ c  c  c c } 
    \toprule
    Split & CIFAR-100 & STL-10 & Mem. (MB) \\ 
    \midrule
   E2E & 51.78 & 59.93 &  37728   \\ 
    K=2 (Ours) & 47.69 &  60.12 & 21200 \color{red} $(\downarrow 43.81\%)$ \\ 
    K=4 (Ours) & 54.31 &  55.09 & 12229 \color{red} $(\downarrow 66.59\%)$ \\ 
    \bottomrule
    \end{tabular}
\caption{Classification accuracy comparison for split $K=2$ and $K=4$ on CIFAR-100 and STL-10, using our proposed local learning on Deconv. decoder. Input resolution is $384 \times 384$ with a batch size of 40, trained for 100 epochs.}
\label{tab:Base_model}
\end{center}
\end{table}

\begin{table}
\begin{center}
    \begin{tabular}{ c   c  c  c } 
    \toprule
    Split & CIFAR-100 & STL-10 & Mem. (MB) \\ 
    %
    \midrule
   E2E & 53.31 & 70.216 & 29851  \\ 
    K=2 (Ours) & 69.72  &  65.52 & 19311 \color{red} $(\downarrow 35.31\%)$ \\ 
    K=4 (Ours) & 72.78  &  63.09 & 11907 \color{red} $(\downarrow 60.11\%)$\\ 
    \bottomrule
    \end{tabular}
\caption{Swin-Large Accuracy comparison for CIFAR-100 and STL-10, using our proposed local learning on Deconv. decoder. Input resolution is $224 \times 224$ with a batch size of 80, trained for 200 epochs.}
\label{tab:large_model}
\end{center}
\end{table}

\section{Conclusion}
In this paper, we propose a novel local learning framework on Transformers. 
Different from previous approach of reconstructing entire images in each local module, our method reconstructs the features from the previous module to improve the performance and save GPU memory.
Our method has been tested on four commonly used classification datasets, proving its superior performance gain over previous method and superior memory saving over end-to-end training pipeline.
In particular, we evaluated our method using three commonly used decoder structures and four datasets, CIFAR-10, CIFAR-100, STL-10 and SVHN. 
It reported on Swin-Tiny, our approach outperforms InfoPro-Transformer by up to $0.58\%$ on these datasets, while using up to $12\%$ less memory. 
Compared to the end-to-end (E2E) training strategy, our method requires $36\%$ less GPU memory when the network is divided into 2 modules and requires $45\%$ less GPU memory when the network is divided into 4 modules, at the cost of little or no accuracy drop. 
Our approach is further proved to work for other variations of the Swin Transformer, namely the Swin-Large and Swin-Base Transformers \cite{swin}.

{\small
\bibliographystyle{ieee_fullname}
\bibliography{egbib}
}

\end{document}